\documentclass[journal]{IEEEtran}

\ifCLASSINFOpdf

\else

\fi
\usepackage[english]{babel}
\usepackage[utf8x]{inputenc}
\usepackage{amsmath}
\usepackage{amsthm}
\usepackage{graphicx}
\usepackage[colorinlistoftodos]{todonotes}
\usepackage{mathtools}
\usepackage{subcaption}
\usepackage{amssymb}
\usepackage[export]{adjustbox}
\usepackage{stfloats}
\usepackage{lettrine}
\usepackage{tikz}
\usetikzlibrary{calc}

\newtheorem{theorem}{Theorem}
\newtheorem{lemma}[theorem]{Lemma}

\newcommand{\tikzmark}[1]{\tikz[overlay,remember picture] \node (#1){};}

\begin{document}
\title{Over-the-Air Federated Learning In Broadband Communication}

\author{Wayne Lemieux,~\IEEEmembership{Fellow,~IEEE,} Raphael Pinard,~\IEEEmembership{Student Member,~IEEE}
        , Mitra Hassani}

%

\markboth{}%
{Shell \MakeLowercase{\textit{et al.}}: Bare Demo of IEEEtran.cls for IEEE Journals}

\maketitle

\begin{abstract}
Federated learning (FL) is a privacy-preserving distributed machine learning paradigm that operates at the wireless edge. It enables clients to collaborate on model training while keeping their data private from adversaries and the central server. However, current FL approaches have limitations. Some rely on secure multiparty computation, which can be vulnerable to inference attacks. Others employ differential privacy, but this may lead to decreased test accuracy when dealing with a large number of parties contributing small amounts of data. To address these issues, this paper proposes a novel approach that integrates federated learning seamlessly into the inner workings of MIMO (Multiple-Input Multiple-Output) systems.
\end{abstract}


\IEEEpeerreviewmaketitle

\section{Introduction}
Channel capacity refers to the maximum data transmission rate achievable with an extremely low probability of errors \cite{gallager1968information}..
Shannon (1948) first tackled the capacity of a single-input-single-output (SISO) additive white Gaussian noise (AWGN) channel.
The introduction of powerful space-time coding schemes (\cite{shannon1948mathematical} has unveiled the potential of multiple-input-multiple-output (MIMO) systems to surpass traditional techniques, presenting academia and industry with a means to achieve significantly higher channel capacity.
This innovative approach provides a unique solution to meet the escalating demand for high-performance next-generation wireless communications.
\\
In this paper, MIMO satellite communication (SatCom) is considered where the receiver is a linear array antenna. We aim to find the channel capacity for low Earth orbiting satellites whose positions are unknown to the terrestrial receiver. In low orbiting satellites, line of sight (LOS) becomes more dominant and the path loss reduces. Although its efficiency in a rich scattering environment has already been
demonstrated \cite{corazza2007digital}, less is studied under this new scenario. 
\\
The typical way to analyze channel capacity of MIMO systems is to find eigenvalue distribution for the channel matrix multiplied by its conjugate \cite{liang2005ergodic}. In our setup, when the channel is modeled as pure LoS, the channel matrix $H$ would become a Vandermonde matrix \cite{horn2012matrix}. However, for this setup, eigenvalue distribution is unknown and only there have been some studies over asymptotic behavior of it \cite{tucci2011eigenvalue,hamidi2022over,hamidi2019systems}. In \cite{tucci2011eigenvalue,struhsaker2020methods}, the author found a lower and upper bound for the maximum eigenvalue and presented the channel capacity for a sufficiently large matrices. 
In this paper, the average channel capacity and outage probability of such channels are analyzed, assuming the receiver has the perfect channel state information (CSI). This has been done by approximating the channel capacity and the accuracy of our method is justified with simulations.  
\\
Federated learning in the context of MIMO (Multiple-Input Multiple-Output) channels refers to a collaborative learning approach where multiple wireless devices jointly train a machine learning model without directly exchanging their raw data.

In federated learning, the MIMO channel acts as the communication medium between the wireless devices and a central server. Instead of transmitting their raw data to the server, the wireless devices locally compute model updates using their own data samples. These updates are then shared with the server over the MIMO channel. The server aggregates the updates from multiple devices to improve the global model without accessing the individual data from each device.

The MIMO channel plays a crucial role in federated learning by enabling efficient and reliable communication between the wireless devices and the central server. The use of multiple antennas in MIMO systems allows for increased data throughput and improved channel capacity, facilitating faster and more reliable transmission of model updates in federated learning scenarios \cite{hamidi2019systems}.

Overall, federated learning in MIMO channels combines the benefits of collaborative machine learning and the capabilities of MIMO systems to enable privacy-preserving and distributed model training across a network of wireless devices \cite{hamidi2022over}.

\section{Channel outage capacity} \label{sec:outage}
Since H is a random matrix, the channel capacity is a random variable. This means that regardless of the rate threshold, there is a non-zero probability that the channel cannot satisfy low error rates \cite{telatar1999capacity}. To describe such channels, the capacity complementary cumulative distribution function is used, known as ccdf \cite{foschini1998limits}, \cite{smith2002gaussian}. This number shows the probability of achieving a specified channel capacity, denoted by $P_{ccdf}$. However, the outage capacity probability is the probability of not achieving a threshold capacity and it is defined as follows
\begin{align} \label{eq:33}
P_{out}= prob \{C<R_{th}\}
\end{align}
where $prob \{ . \}$ is taken over all realization of $H$. It is worth mentioning to say that $P_{out}=1-P_{ccdf}$.
To find the channel capacity the following conjecture is made and left without proof. However, with many simulations, the accuracy of it has been proved to the authors.
\begin{theorem} \label{theorem2}
MIMO channel capacity of the channel described in this paper can be approximated by
Gaussian approximation, for the case when the receiver has the perfect CSI but the transmitter does not
\end{theorem}
This means that to find the outage probability we only need to find $\mathbf{E} \{C \}$ and $\mathbf{Var}  \{C \}$.
\\
To prove the theorem \ref{theorem2}, we refer back to \eqref{eq:14}. Based on the intuition we have obtained so far, the behavior of the channel capacity is mainly dependent on the first three moments of $Trace\{\mathbf{W}\}$. The first moment is a constant number, and therefore the second and the third moments play a substantial role in determining the distribution of channel capacity. 
\\
\begin{lemma}
The distribution of $Trace\{\mathbf{W^2}\}$ converges to the Normal distribution when $n_T$ is sufficiently large.
\end{lemma}
\begin{proof}
What determines the distribution of $Trace\{\mathbf{W^2}\}$ is the second term of equation \eqref{eq:22} which could be re-written as 
\begin{align}
\Omega=\sum_{i=1}^{n_T} \sum_{j=1}^{n_T}\sum_{s=1}^{n_R-1} (n_R-s)\cos(skd\gamma_{ij})
\end{align}
in which the coefficient behind the sum is eliminated (with respect to equation \eqref{eq:22}). 
\\
The summation over $i$ and $j$ in $\Omega$ could be scrambled and replaced by a single summation over a new variable $l$. Also, we define new functions $F_{l}$ , $1 \leq l \leq n_T$, as shown in. By this definition we have
\begin{align}
\Omega=\sum_{l=1}^{n_T} F_{l}
\end{align}

\begin{figure*}[!t]
\normalsize
\begin{equation}
\label{F1}
\scriptstyle{F_1=\underbrace{\sum_{s=1}^{n_R-1} (n_R-s)\cos(skd\gamma_{11})}_{F_{11}}+\underbrace{\sum_{s=1}^{n_R-1} (n_R-s)\cos(skd\gamma_{22})}_{F_{12}}+
\underbrace{\sum_{s=1}^{n_R-1} (n_R-s)\cos(skd\gamma_{33})}_{F_{13}}+...+\underbrace{\sum_{s=1}^{n_R-1} (n_R-s)\cos(skd\gamma_{n_Tn_T})}_{F_{1n_T}}} 
\end{equation}
\begin{equation}
\label{F2}
\scriptstyle{F_2=\underbrace{\sum_{s=1}^{n_R-1} (n_R-s)\cos(skd\gamma_{12})}_{F_{21}}+\underbrace{\sum_{s=1}^{n_R-1} (n_R-s)\cos(skd\gamma_{23})}_{F_{22}}+
\underbrace{\sum_{s=1}^{n_R-1} (n_R-s)\cos(skd\gamma_{34})}_{F_{23}}+...+\underbrace{\sum_{s=1}^{n_R-1} (n_R-s)\cos(skd\gamma_{n_T1})}_{F_{2n_T}}} 
\end{equation}
\begin{equation}
.\nonumber
\end{equation}

\begin{equation}
.\nonumber
\end{equation}

\begin{equation}
.\nonumber
\end{equation}

\begin{equation}
\scriptstyle{F_{n_T}=\underbrace{\sum_{s=1}^{n_R-1} (n_R-s)\cos(skd\gamma_{1n_T})}_{F_{n_T1}}+\underbrace{\sum_{s=1}^{n_R-1} (n_R-s)\cos(skd\gamma_{21})}_{F_{n_T2}}+
\underbrace{\sum_{s=1}^{n_R-1} (n_R-s)\cos(skd\gamma_{32})}_{F_{n_T3}}+...+\underbrace{\sum_{s=1}^{n_R-1} (n_R-s)\cos(skd\gamma_{n_Tn_{T-1}})}_{F_{n_Tn_T}}} 
\end{equation}

\hrulefill
\vspace*{4pt}
\end{figure*}
Now, we note that all $n_T$ terms in each of $F_{l}$'s, where $1 \leq l \leq n_T$, are mutually independent. For instance, we consider $F_{1}$. It is seen that all $F_{1l}$ terms are mutually independent and also have them same distribution; therefore the Central Limit Theorem (CLT) is hold for $F_1$ and so is for other $F_l$'s. Hence, all $F_{l}$'s are normally distributed.  
\\
However, $F_{l}$'s are mutually dependent. Nevertheless, the sum of correlated Gaussian distribution is also Gaussian. Henceforth, we aim to find the parameters for this distribution. 
\\
First, we aim to find the expected value of $\Omega$ denoted by $\mu_{\Omega}$. Due to symmetry, the expected value for all $F_{i,j}$'s, where $1 \leq i \neq j \leq n_T$, are the same. For $F_{12}$, for instance, we have
\begin{align}  \label{E(F12)}
\mathbf{E} (F_{12})=\sum_{s=1}^{n_R-1}(n_R-s) J_0  \big( \frac{skd}{2} \big)^4
\end{align}
\\
Also, for the all $F_{i,j}$'s with $1 \leq i = j \leq n_T$, we have $\cos(skd\gamma_{ii})=1$, and therefore, 
\begin{align}  \label{E(F11)}
\mathbf{E} (F_{11})=\sum_{s=1}^{n_R-1}(n_R-s)= \frac{n_R(n_R-1)}{2}
\end{align}
There are $n_T^2-n_T$ and  $n_T$ terms in $\Omega$ with the same expected value as $F_{12}$ and $F_{11}$, respectively. Therefore, 

\begin{align}  \label{mu_omega}
\mu_{\Omega}=
\end{align}

The correlated terms between $F_{1}$ and $F_2$ are the pairs $(F_{11},F_{21})$, $(F_{12},F_{22})$,..., $(F_{1n_T},F_{2n_T})$ and the other $n_T\times \ (n_T-1)$ terms are mutually independent. Hence we have
\begin{align}  \label{covF}
\mathbf{E}(F_1F_{2})-\mathbf{E}(F_1)\mathbf{E}(F_{2})  ~~~~~~~~\nonumber \\
=\sum_{r=1}^{n_T} \mathbf{E} (F_{1r}F_{2r})-\sum_{r=1}^{n_T} \mathbf{E} (F_{1r})\mathbf{E} (F_{2r})  ~~~~~~~~\nonumber \\
=n_T \times \mathbf{E} (F_{1n_T}F_{2n_T})-n_T \times \mathbf{E} (F_{1n_T})\mathbf{E} (F_{2n_T})
\end{align}
Where the last equation is obtained because of the symmetry. To evaluate equation \eqref{covF}, first we find $\mathbf{E} (F_{1n_T}F_{2n_T})$ and then $\mathbf{E} (F_{1n_T})\mathbf{E} (F_{2n_T})$.

\begin{itemize}
\item{$\mathbf{E} (F_{1n_T}F_{2n_T})$}
\\
First, we evaluate the terms in which coefficient $s$ is common, i.e., $A=\mathbf{E}\big(\cos(skd\gamma_{1n_T}) \cos(skd\gamma_{2n_T})  \big)$, where $s=1,2,...,n_R-1$.
\begin{align}  \label{cos}
\cos(skd\gamma_{1n_T}) \cos(skd\gamma_{2n_T})  ~~~~~~~~\nonumber \\
=\frac{1}{2} \Big( \underbrace{\cos \big( skd(\gamma_{1n_T}+\gamma_{2n_T}) \big)}_{A1}+ \underbrace{\cos \big( skd(\gamma_{1n_T}-\gamma_{2n_T}) \big)}_{A2}  \Big) \nonumber \\
\end{align}
Now, we find $\mathbf{E} (A1)$ and $\mathbf{E} (A2)$ separately. Based on the definition of $\gamma$, we have
\begin{align}  \label{A1}
\mathbf{E} (A1)=\mathbf{E} \bigg( \cos \Big( skd \big( \sin(\theta_1)\sin(\phi_1) +\sin(\theta_2)\sin(\phi_2) \nonumber \\ 
- 2\sin(\theta_{n_T})\sin(\phi_{n_T}) \big) \Big) \bigg) ~~~~~~~~~~~~~\nonumber \\
=\mathbf{E} \bigg( \cos \Big( skd \big( \sin(\theta_1)\sin(\phi_1) +\sin(\theta_2)\sin(\phi_2)\big) \Big) \bigg)\nonumber \\
\times \mathbf{E}\bigg(\cos \Big( skd \big( 2\sin(\theta_{n_T})\sin(\phi_{n_T}) \big) \Big) \bigg) ~~~~~~~~~~~~~\\ \label{eq:sin}
=\mathbf{E}\bigg(\cos \Big( skd\sin(\theta_{1})\sin(\phi_{n1}) \Big) \bigg) ~~~~~~~~~~~~~\nonumber \\
\times \mathbf{E}\bigg(\cos \Big( skd\sin(\theta_{2})\sin(\phi_{n2}) \Big) \bigg) ~~~~~~~~~~~~~\nonumber \\
\times \mathbf{E}\bigg(\cos \Big( 2skd\sin(\theta_{n_T})\sin(\phi_{n_T}) \Big) \bigg) ~~~~~~~~~~~~~\\ \label{eq:j}
=J_0  \big( \frac{skd}{2} \big)^4 \times J_0  \big( skd \big)^2 ~~~~~~~~~~~~~
\end{align}
Where \eqref{eq:sin} is hold because the $\sin(.)$ term in $\cos(.)$ expansion is odd. Also \eqref{eq:j} could be derived as the same as equation \eqref{eq:29} \cite{struhsaker2020methods}.
\\
$\mathbf{E} (A2)$ could be found similarly and for the sake of brevity we only mention the final answer
\begin{align}  \label{A2}
\mathbf{E} (A2)=J_0  \big( \frac{skd}{2} \big)^4
\end{align}
Therefore 
\begin{align}  \label{Ecos}
A=\mathbf{E}\big( \cos(skd\gamma_{1n_T}) \cos(skd\gamma_{2n_T})  \big) \nonumber \\
\frac{1}{2} J_0  \big( \frac{skd}{2} \big)^4 \big( 1+  J_0  \big( skd \big)^2   \big)
\end{align}
Second, we evaluate the terms with different coefficient $s$ and we call them $s_ 1$ and $s_2$. As equation \eqref{cos} we have: 
\begin{align}  \label{cos2}
\cos(s_1kd\gamma_{1n_T}) \cos(s_2kd\gamma_{2n_T})  ~~~~~~~~~~~~~~~~~~\nonumber \\
=\frac{1}{2} \Big( \underbrace{\cos \big( s_1kd\gamma_{1n_T}+s_2kd\gamma_{2n_T} \big)}_{B1}+ \underbrace{\cos \big( s_1kd\gamma_{1n_T}-s_2kd\gamma_{2n_T} \big)}_{B2}  \Big) \nonumber \\
\end{align}
Similar to  $\mathbf{E} (A1)$, one can find $\mathbf{E} (B1)$ as follows
\begin{align}  \label{B1}
\mathbf{E} (B1)
=J_0  \big( \frac{s_1kd}{2} \big)^2 J_0  \big( \frac{s_2kd}{2} \big)^2 J_0  \big( \frac{(s_1+s_2)kd}{2} \big)^2 
\end{align}
Also, 
\begin{align}  \label{B2}
\mathbf{E} (B2)
=J_0  \big( \frac{s_1kd}{2} \big)^2 J_0  \big( \frac{s_2kd}{2} \big)^2 J_0  \big( \frac{(s_1-s_2)kd}{2} \big)^2 
\end{align}
Therefore, 

\begin{align}  \label{B1+B2}
B=\mathbf{E} \big( \cos(s_1kd\gamma_{1n_T}) \cos(s_2kd\gamma_{2n_T}) \big) ~~~~~~~~~~~~~~~~~~\nonumber \\
=\frac{J_0  \big( \frac{s_1kd}{2} \big)^2 J_0  \big( \frac{s_2kd}{2} \big)^2}{2} \Big(J_0 \big( \frac{(s_1+s_2)kd}{2} \big)^2 + J_0 \big( \frac{(s_1-s_2)kd}{2} \big)^2 \Big)
\end{align}

Now, we return back to find $\mathbf{E} (F_{1n_T}F_{2n_T})$. In this term, there are some $A$ and $B$ type terms each of which appearing with different coefficients. We break down the terms in $ F_{1n_T}$ and $F_{2n_T}$ as shown in table \ref{table:1}. $F_{1n_T}$ and $F_{2n_T}$ and their corresponding terms are written in each column. In this table, the red and blue arrows show expected value of type A and B, respectively.

\begin{table}[h!]
   \begin{tabular}[t]{ll}
   $\mathbf{F_{1n_T}}$  & ~~~~~~~~~~$ \mathbf{F_{2n_T}}$   \\ \hline \hline
$(n_R-1)\cos(kd\gamma_{1n_T})$~\tikzmark{11} &~~~~~~~~~~$\tikzmark{12} (n_R-1)\cos(kd\gamma_{2n_T})$  \\ 
$(n_R-2)\cos(2kd\gamma_{1n_T})$\tikzmark{21}&~~~~~~~~~~$\tikzmark{22} (n_R-2)\cos(2kd\gamma_{2n_T})$  \\ 
.&~~~~~~~~~~. \\
.&~~~~~~~~~~.\\
.&~~~~~~~~~~.\\
$\cos((n_R-1)kd\gamma_{1n_T})$~\tikzmark{L1}&~~~~~~~~~~$\tikzmark{L2} \cos((n_R-1)kd\gamma_{2n_T})$  \\ 
   \end{tabular}
\caption{The terms in $ F_{1n_T}$ and $F_{2n_T}$. The red and blue arrows show expected value of type A and B, respectively}
\label{table:1}
\end{table}

\tikz[overlay,remember picture]\draw[red,<->] ($(11)$)+(.1em,0.2em)--($(12)+(-.6em,0.2em)$);
\tikz[overlay,remember picture]\draw[red,<->] ($(21)$)+(.1em,0.2em)--($(22)+(-.6em,0.2em)$);
\tikz[overlay,remember picture]\draw[red,<->] ($(L1)$)+(.1em,0.2em)--($(L2)+(-.6em,0.2em)$);

\tikz[overlay,remember picture]\draw[blue,<->] ($(11)+(.1em,0.2em)$)--($(22)+(-.6em,0.2em)$);
\tikz[overlay,remember picture]\draw[blue,<->] ($(11)+(.1em,0.2em)$)--($(L2)+(-.6em,0.2em)$);

\tikz[overlay,remember picture]\draw[blue,<->] ($(21)+(.1em,0.2em)$)--($(12)+(-.6em,0.2em)$);
\tikz[overlay,remember picture]\draw[blue,<->] ($(21)+(.1em,0.2em)$)--($(L2)+(-.6em,0.2em)$);

\tikz[overlay,remember picture]\draw[blue,<->] ($(L1)+(.1em,0.2em)$)--($(12)+(-.6em,0.2em)$);
\tikz[overlay,remember picture]\draw[blue,<->] ($(L1)+(.1em,0.2em)$)--($(22)+(-.6em,0.2em)$);

Hence, 

\begin{align}  \label{red}
\mathbf{E} (F_{1n_T}F_{2n_T})=\mathbf{E} \{ Red ~Arrows\} + \mathbf{E} \{ Blue~ Arrows\} 
\end{align}
Summation over red arrows, which are expected value of type A, yields to

\begin{align}  \label{red}
\mathbf{E} \{ Red ~Arrows\}=\sum_{s=1}^{n_R-1} (n_R-s)^2 \nonumber \\
\times \frac{1}{2} J_0  \big( \frac{skd}{2} \big)^4 \big( 1+  J_0  \big( skd \big)^2   \big)
\end{align}
Also for the blue arrows we have
\begin{align}  \label{blue}
\mathbf{E} \{ Blue ~Arrows\}=\sum_{s_2=1}^{n_R-1} \sum_{s_1\neq s_2}^{n_R-1} (n_R-s_2)(n_R-s_1) \nonumber \\
\times \frac{J_0  \big( \frac{s_1kd}{2} \big)^2 J_0  \big( \frac{s_2kd}{2} \big)^2}{2}~~~~~~~~~~~ \nonumber \\
\times \Big(J_0 \big( \frac{(s_1+s_2)kd}{2} \big)^2 + J_0 \big( \frac{|s_1-s_2|kd}{2} \big)^2 \Big)
\end{align}

\item{$\mathbf{E} (F_{1n_T})\mathbf{E} (F_{2n_T})$}
\\
Due to symmetry we have $\mathbf{E} (F_{1n_T})=\mathbf{E} (F_{2n_T})$ and
\begin{align}  \label{E(F1)E(F2)}
\mathbf{E} (F_{1n_T})=\sum_{s=1}^{n_R-1}(n_R-s) J_0  \big( \frac{skd}{2} \big)^4
\end{align}
\end{itemize}

Now, we aim to find the denominator of equation \eqref{eq:CF}. First we note that, due to symmetry, $\sqrt {\mathbf{Var}(F_1)\mathbf{Var}(F_{n+1})}=\mathbf{Var}(F_{1})$.
\\
Also,
\begin{align}  \label{Varcov}
\mathbf{Var}(F_{1})=\sum_{i=1}^{n_T} \mathbf{Var}(F_{1i})+ \sum_{i=1}^{n_T} \sum_{i\neq j}^{n_T} \mathbf{Cov}(F_{1i}F_{1j}) \\
=n_T\mathbf{Var}(F_{1i})+\frac{n_T(n_T-1)}{2}\mathbf{Cov}(F_{1n_T}F_{1n_{T}-1})
\end{align}
Where the latter is obtained due to symmetry. $\mathbf{Cov}(F_{1n_T}F_{1n_{T}-1})$ was implicitly obtained in the previous part. 
For $\mathbf{Var}(F_{1n_T})$, we need to first find 
\begin{align}  \label{Var}
\mathbf{E}(F_{1n_T})=\sum_{s=1}^{n_R-1}(n_R-s) J_0  \big( \frac{skd}{2} \big)^4
\end{align}
Furthermore, 
\begin{align} 
\mathbf{Var}(F_{1n_T})=\sum_{s=1}^{n_R-1} (n_R-s)^2 
\times \frac{1}{2} \big( 1+  J_0  \big( skd \big)^4   \big) \nonumber \\
+\sum_{s_2=1}^{n_R-1} \sum_{s_1\neq s_2}^{n_R-1} (n_R-s_2)(n_R-s_1) \nonumber \\
\times \frac{1}{2}\Big(J_0 \big( \frac{(s_1+s_2)kd}{2} \big)^4 + J_0 \big( \frac{|s_1-s_2|kd}{2} \big)^4 \Big) \nonumber \\
-\Big( \sum_{s=1}^{n_R-1}(n_R-s) J_0  \big( \frac{skd}{2} \big)^4 \Big)^2
\end{align}
All in all, correlation coefficient could be simplified as stated in equation 

\begin{figure*}[!t]
\normalsize

\begin{equation}
\scriptstyle {C_F(n)=\frac{\mathbf{Cov}(F_{1n_T}F_{1n_{T}-1})}{(\frac{n_T-1}{2})\mathbf{Cov}(F_{1n_T}F_{1n_{T}-1})+\sum_{s_2=1}^{n_R-1} \sum_{s_1=1}^{n_R-1} (n_R-s_2)(n_R-s_1) \frac{1}{2}\Big(J_0 \big( \frac{(s_1+s_2)kd}{2} \big)^4 + J_0 \big( \frac{|s_1-s_2|kd}{2} \big)^4 \Big)-\sum_{s_2=1}^{n_R-1}\sum_{s_1=1}^{n_R-1} (n_R-s_2)(n_R-s_1) J_0  \big( \frac{s1kd}{2} \big)^4 J_0  \big( \frac{s2kd}{2} \big)^4}}
\end{equation}
\begin{equation}
\label{CFmain}
\scriptstyle {C_F(n)=\frac{\sum_{s_2=1}^{n_R-1} \sum_{s_1=1}^{n_R-1} (n_R-s_2)(n_R-s_1)
\frac{J_0  \big( \frac{s_1kd}{2} \big)^2 J_0  \big( \frac{s_2kd}{2} \big)^2}{2}
\Big(J_0 \big( \frac{(s_1+s_2)kd}{2} \big)^2 + J_0 \big( \frac{|s_1-s_2|kd}{2} \big)^2 \Big)-\Big( \sum_{s=1}^{n_R-1}(n_R-s) J_0  \big( \frac{skd}{2} \big)^4 \Big)^2}
{\sum_{s_2=1}^{n_R-1} \sum_{s_1=1}^{n_R-1} (n_R-s_2)(n_R-s_1) \frac{1}{2}\Big(J_0 \big( \frac{(s_1+s_2)kd}{2} \big)^4 + J_0 \big( \frac{|s_1-s_2|kd}{2} \big)^4 \Big)}}
\end{equation}
\hrulefill
\vspace*{4pt}
\end{figure*}

\end{proof}
The remaining of this section is divided into 2 subsections; In the \ref{variance}, we find the variance of the channel and in \ref{simout}, $P_{out}$ is calculated and some simulation results are provided to demonstrate the accuracy of the conjecture we made. 

\subsection{Variance of the channel capacity} \label{variance}
Since $\mathbf{Var}  \{C \}=\mathbf{E} \{C^2 \}- \big( \mathbf{E}\{C \} \big)^2$ and we have already found $\mathbf{E} \{C \}$ in \ref{E(C)}, we aim to find $\mathbf{E} \{C^2 \}$ in this part. 
\\
We have
\begin{align}  \label{eq:34}
C^2=\Big( \sum_{i=1}^{n_T}  \log_2 \big( 1+\frac{P}{\sigma^2} \lambda_i \big) \Big)^2~~~~~~~~~~~~~~~~~~ \nonumber \\
=\underbrace {\sum_{i=1}^{n_T}  \log_2^2 \big( 1+\frac{P}{\sigma^2} \lambda_i \big)}_{C1}+\underbrace {\sum_{i\neq j}^{n_T}\log_2 \big( 1+\frac{P}{\sigma^2} \lambda_i \big) \log_2 \big( 1+\frac{P}{\sigma^2} \lambda_j \big)}_{C2}
\end{align}
and each $\log_2(.)$ term can be approximated by its Taylor expansion. The results we obtained in section \ref{E(C)} suggest that we can only keep those terms with their power less than equal to 3 to approximate the $\log$ terms. 
\\
First we find $C1$
\begin{align}  \label{eq:35}
C1=\big( \frac{1}{\ln(2)}\big)^2 \sum_{i=1}^{n_T} \Big(\frac{P}{\sigma^2} \lambda_i  -\frac{(\frac{P}{\sigma^2} \lambda_i)^2}{2}+\frac{(\frac{P}{\sigma^2} \lambda_i)^3}{3}+... \Big)^2 \nonumber \\
=\big( \frac{1}{\ln(2)}\big)^2\sum_{i=1}^{n_T} \Big(\big( \frac{P}{\sigma^2} \lambda_i \big)^2-\big( \frac{P}{\sigma^2} \lambda_i \big)^3+...\Big) \nonumber \\
\approx \big( \frac{1}{\ln(2)}\big)^2 \Big( \big( \frac{P}{\sigma^2}\big)^2 Trace(\mathbf{W}^2)-\big( \frac{P}{\sigma^2}\big)^3 Trace(\mathbf{W}^3) \Big)
\end{align}
For $C2$ we have
\begin{align}  \label{eq:36}
C2=\big( \frac{1}{\ln(2)}\big)^2~~~~~~~~~~~~~~~~~~~~~~~~~~~~~~~~~~~~~~~~~~~~~~~~ \nonumber \\
\times \sum_{i\neq j}^{n_T} \Big(\frac{P}{\sigma^2} \lambda_i  -\frac{(\frac{P}{\sigma^2} \lambda_i)^2}{2}+...\Big) \Big(\frac{P}{\sigma^2} \lambda_j  -\frac{(\frac{P}{\sigma^2} \lambda_j)^2}{2}+...\Big)  ~~~ ~~~~~\nonumber \\
\approx \big( \frac{1}{\ln(2)}\big)^2 \sum_{i\neq j}^{n_T} \Big(\big( \frac{P}{\sigma^2}\big)^2 \lambda_i  \lambda_j - \big( \frac{P}{\sigma^2}\big)^3 \frac{\lambda_i^2 \lambda_j}{2}- \big( \frac{P}{\sigma^2}\big)^3 \frac{\lambda_i \lambda_j^2}{2}\Big)~~~~~\nonumber \\
=\big( \frac{1}{\ln(2)}\big)^2 \Big( \underbrace { \sum_{i\neq j}^{n_T} \big( \frac{P}{\sigma^2}\big)^2 \lambda_i  \lambda_j}_{C21} - \underbrace { \sum_{i\neq j}^{n_T}  \big( \frac{P}{\sigma^2}\big)^3 \lambda_i^2}_{C22} \lambda_j \Big)~~~~~~~~~~
\end{align}
To find $C21$ we know that
\begin{align}  \label{eq:37}
\big(Trace(\mathbf{W})\big)^2=\Big( \sum_{i=1}^{n_T} \big(\lambda_i \big) \Big)^2  \nonumber \\
= \sum_{i=1}^{n_T} \big(\lambda_i^2 \big)+  \sum_{i\neq j}^{n_T} \lambda_i  \lambda_j  \nonumber \\ 
=Trace(\mathbf{W^2})+ \sum_{i\neq j}^{n_T} \lambda_i  \lambda_j 
\end{align}
and therefore
\begin{align}  \label{eq:38}
C21=\big( \frac{P}{\sigma^2}\big)^2 \Big(\big(Trace(\mathbf{W})\big)^2- Trace(\mathbf{W^2}) \Big)
\end{align}
To find $C22$, considering that $\sum_{i=1}^{n_T} \big(\lambda_i \big)=Trace(\mathbf{W})=1$, we have
\begin{align}  \label{eq:39}
Trace(\mathbf{W^2})=\Big( \sum_{i=1}^{n_T} \lambda_i^2 \Big) \Big(  \sum_{i=1}^{n_T} \lambda_i \Big) \nonumber \\ 
= \sum_{i=1}^{n_T} \lambda_i^3 +\sum_{i\neq j}^{n_T} \lambda_i^2  \lambda_j  \nonumber \\ 
=Trace(\mathbf{W^3})+\sum_{i\neq j}^{n_T} \lambda_i^2  \lambda_j
\end{align}
and thus
\begin{align}  \label{eq:40}
C22=\big( \frac{P}{\sigma^2}\big)^3 \Big(\big(Trace(\mathbf{W})\big)^2- Trace(\mathbf{W^3}) \Big)
\end{align}
Now we can find $C^2$ as follows:
\begin{align}  \label{eq:41}
C^2 \approx \big( \frac{1}{\ln(2)}\big)^2 \Big( \big( Trace(\mathbf{W})\big)^2-Trace(\mathbf{W^2})\Big) \nonumber \\
=\big( \frac{1}{\ln(2)}\big)^2 \Big(1-Trace(\mathbf{W^2})\Big)~~~~~~~~
\end{align}
Hence, 
\begin{align}  \label{eq:42}
\mathbf{E} \{C^2 \} \approx \big( \frac{1}{\ln(2)}\big)^2  \Big(1-\mathbf{E} \big\{Trace(\mathbf{W^2})\big\}\Big)
\end{align}
and $\mathbf{E} \big\{Trace(\mathbf{W^2})\big\}$ is found by equation \eqref{eq:30}
\subsection{Calculating $P_{out}$ and demonstrating its accuracy} \label{simout}
Now, considering equation \eqref{eq:33} the Gaussian approximated outage capacity $P_{out}$ can be found as follows
\begin{align}  \label{eq:43}
P_{out}=1-Q \Big( \frac{R_{th}-\mathbf{E}\{C\}}{\sqrt{\mathbf{Var}  \{C \}}} \Big) \nonumber \\
=Q \Big( \frac{\mathbf{E}\{C\}-R_{th}}{\sqrt{\mathbf{Var}  \{C \}}} \Big)
\end{align}
where $Q$ is the tail distribution function of the standard normal distribution and is defined as
\begin{align}  \label{eq:44}
Q(x)=\frac{1}{\sqrt{2\pi}}\int_{x}^{\infty} e^{\frac{-t^2}{2}}dt 
\end{align}
Here, we show the simulation result for outage capacity and we compare it to the one found by \eqref{eq:43}.
\begin{figure}[htbp] \label{fig:outage}
\centering{\includegraphics[scale=0.4]{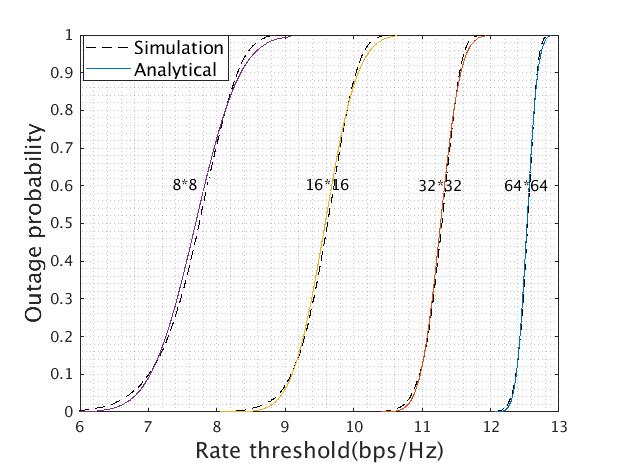}}
   \caption{Outage capacity for different number of satellites and receive antenna elements, SNR=10dB. Solid lines and dashed lines represent the outage capacity found by equation \eqref{eq:43} and simulation, respectively.}
   \label{fig:outage}
\end{figure}
Figure \ref{fig:outage} justifies the accuracy of our conjecture.

\begin{figure}[h!] \label{fig:outage2}
\includegraphics[width=.65\textwidth,center]{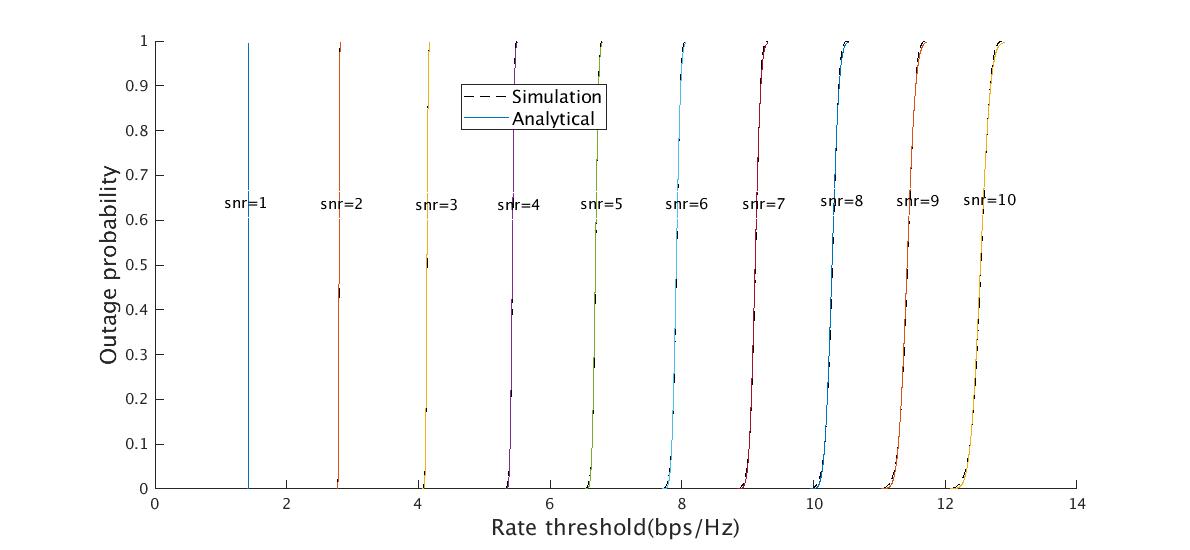}
   \caption{Outage capacity of a 64 $\times$ 64 MIMO setup for different SNR's. Solid lines and dashed lines represent the outage capacity found by equation \eqref{eq:43} and simulation, respectively.}
   \label{fig:outage2}
\end{figure}

\appendices
\section{Array factor derivation}\label{app:AF}
The more general form the array factor expression is found using the following expression:
\begin{align} \label{eq:AF}
AF(\theta, \phi)=\sum_{m=0}^{M-1} I_m e^{jk\hat{r}.\overrightarrow{r}_m}
\end{align}
where $\overrightarrow{r}_m$ is the position vector to the $m^{th}$ element and $\hat{r}$ is a unit vector pointing in the direction of interest, i.e.,
\begin{align}\label{eq:rhat}
\hat{r}=\sin\theta \cos\phi \hat{x}+\sin\theta \sin\phi \hat{y}+\cos\theta \hat{z}
\end{align}
and for a linear configuration evenly distributed along y axis with spacing d, $\overrightarrow{r}_m=md \hat{y}$. Therefore $\hat{r}.\overrightarrow{r}_m=\sin\theta \sin\phi$ and the equation \ref{eq:4} is derived.

\section{Proof of theorem \label{th:lambda}}
Characteristic polynomial of an $n\times n$matrix $A$ could be written in terms of traces of matrices $A, A^2, ..., A^n$ as follows \cite{pennisi1987coefficients}
\begin{align}  \label{eq:52}
P(\lambda)=b_0\lambda^n+b_1\lambda^{n-1}+b_2\lambda^{n-2}+ ...+b_{n-1}\lambda+b_n
\end{align}
where 
\begin{align} 
b_0=(-1)^n, ~~~~~~ b1=-(-1)^n \mathbf{T}_1^A ~~~~~~~~~~~~  \nonumber \\
b_2=-\frac{1}{2} \big( b_1 \mathbf{T}_1^A + (-1)^n \mathbf{T}_2^A \big)~~~~~~~~~~~~ \nonumber \\
. ~~~~~~~~~~~~~~~~~~~~~~~~ \nonumber \\
. ~~~~~~~~~~~~~~~~~~~~~~~~ \nonumber \\
. ~~~~~~~~~~~~~~~~~~~~~~~~ \nonumber \\
b_n=-\frac{1}{n}  \big( b_{n-1} \mathbf{T}_1^A +b_{n-2} \mathbf{T}_2^A +...+b_{1} \mathbf{T}_{n-1}+ (-1)^n \mathbf{T}_n^A \big)
\end{align}
and $\mathbf{T}_1^A, \mathbf{T}_2^A, ..., \mathbf{T}_n^A$ denote the traces of matrices $A, A^2, ..., A^n$.
Also, the summation of roots of equation \eqref{eq:52} is equal to $\frac{-b_1}{b_0}=\mathbf{T}_1^A$.
\\
On the other hand, eigenvalues of matrix $A^k$ are  eigenvalues of matrix $A$ each to the power of $k$.
Therefore if $\lambda_1, \lambda_2,...,\lambda_n$ are eigenvalues of matrix $A$, then $\lambda_1^k, \lambda_2^k,...,\lambda_n^k$ would be those of matrix $A^k$, and thus
\begin{align}  \label{eq:52}
\sum_{i=1}^n \lambda_i^k=\mathbf{T}_1^{A^k}
\end{align}
And the theorem is proved.

\section{Average channel capacity for a linear array along z axis}\label{Z axis}
The array factor for a linear array along z axis is mentioned in \eqref{eq:49}, where $0 \leq \theta \leq \frac{\pi}{2}$. It stands to reason that again $Trace(\mathbf{W})=1$. Using the formula we obtained in \eqref{eq:23}, one can justify that
\begin{align} 
\mathbf{E} \Big\{Trace(\mathbf{W}^2) \Big\}=\frac{1}{(n_Rn_T)^2}\Big(n_R^2n_T+n_Rn_T(n_T-1)+\nonumber ~~~~~~~~~~~~~\\
2\sum_{s=1}^{n_R-1}(n_R-s)n_T(n_T-1) J_0  \big( \frac{skd}{2} \big)^2\Big)~~~~~~~~~~~~~
\end{align}
Also to obtain the third moment from equation \eqref{eq:31}, we have
\begin{flalign}
\mathbf{E} \Big\{Trace(\mathbf{W}^2) \Big\}= \frac{1}{(n_Rn_T)^3} \Bigg(n_R^3nT+3n_R^2n_T(n_T-1)~~~~~~ \nonumber \\
+n_Rn_T(n_T-1)(n_T-2)~~~~~~ ~~~~~~ \nonumber \\
+6\sum_{s=1}^{n_R-1}n_R(n_R-s)n_T(n_T-1) J_{0} \big( \frac{skd}{2} \big)^4~~~~~~ ~~~~~~ \nonumber \\+6\sum_{s=1}^{n_R-1}(n_T-2)\prod_{k=0}^2(k+s)J_{0} \big( \frac{(n_R-s)kd}{2} \big)^2~~~~~ ~~~\nonumber \\
+6\sum_{s=1}^{\frac{n_R-1}{2}}(n_T-2)\prod_{k=0}^2(n_R-2s+k)J_{0} \big( \frac{skd}{2} \big)^4 J_{0} \big( {skd} \big)~~~~~ ~~~\nonumber \\
+\Big(6\sum_{s=1}^{n_R-1}\sum_{t=1}^{n_R-2s} (n_R-2m)^{+}(n_T-2) \times~~~~~ ~~~ \nonumber \\
\prod_{k=0}^2 (2n_R-2t-2s+2+k) \times ~~~~~ ~~~\nonumber \\
J_{0} \big( \frac{skd}{2} \big) J_{0} \big( \frac{(s+t)kd}{2} \big)J_{0} \big( \frac{(2s+t)kd}{2} \big) \Big)\Bigg)~~~~ ~~~~~
\end{flalign}
\bibliographystyle{IEEEtran}
\bibliography{refs}
\end{document}